\documentclass{mva_style}
\usepackage{graphicx}
\usepackage[dvipsnames]{xcolor}
\usepackage{amsmath}
\usepackage{amssymb}
\usepackage{mathtools}
\usepackage{booktabs}
\usepackage{xspace}
\usepackage{here}
\usepackage{pifont}

\makeatletter
\let\MYcaption\@makecaption
\makeatother
\usepackage[subrefformat=parens]{subcaption}
\makeatletter
\let\@makecaption\MYcaption
\makeatother

\usepackage[breaklinks,colorlinks]{hyperref}

\usepackage[capitalize]{cleveref}
\crefname{section}{Sec.}{Secs.}
\Crefname{section}{Section}{Sections}
\crefname{table}{Tab.}{Tabs.}
\Crefname{table}{Table}{Tables}
\crefname{figure}{Fig.}{Figs.}
\Crefname{figure}{Figure}{Figures}
\crefname{equation}{Eq.}{Eqs.}

\finalcopy %

\makeatletter
\DeclareRobustCommand\onedot{\futurelet\@let@token\@onedot}
\def\@onedot{\ifx\@let@token.\else.\null\fi\xspace}

\def\eg{\emph{e.g}\onedot}

\def\etal{\emph{et al}\onedot}
\makeatother

\newcommand{\TP}{\mathrm{TP}}
\newcommand{\FP}{\mathrm{FP}}
\newcommand{\FN}{\mathrm{FN}}
\newcommand{\wTP}{\mathrm{wTP}}
\newcommand{\wFP}{\mathrm{wFP}}
\newcommand{\wFN}{\mathrm{wFN}}
\newcommand{\Recall}{\mathrm{Recall}}
\newcommand{\Precision}{\mathrm{Precision}}
\newcommand{\wRecall}{\mathrm{wRecall}}
\newcommand{\wPrecision}{\mathrm{wPrecision}}

\newcommand{\M}{\,M\xspace}

\newcommand{\MSEC}{\,ms\xspace}

\newcommand{\cm}{\textcolor{Green}{\ding{51}}\xspace}
\newcommand{\xm}{\textcolor{Red}{\ding{55}}\xspace}

\newcommand{\APS}{AP$_\textit{S}$\xspace}
\newcommand{\APM}{AP$_\textit{M}$\xspace}
\newcommand{\APL}{AP$_\textit{L}$\xspace}

\AtBeginDocument{
	\abovedisplayskip     =0.3\abovedisplayskip
	\abovedisplayshortskip=0.3\abovedisplayshortskip
	\belowdisplayskip     =0.3\belowdisplayskip
	\belowdisplayshortskip=0.3\belowdisplayshortskip
}

\begin{document}
\title{BandRe: Rethinking Band-Pass Filters\\for Scale-Wise Object Detection Evaluation}

\author{
Yosuke Shinya\thanks{This work was done independently of the author's employer.}\\
Independent researcher\\
Tokyo, Japan\\
{\tt \url{https://shinya7y.github.io/}}\\
}

\maketitle

\section*{\centering Abstract}
\textit{
	Scale-wise evaluation of object detectors is important for real-world applications.
	However, existing metrics are either coarse or not sufficiently reliable.
	In this paper, we propose novel scale-wise metrics that strike a balance between fineness and reliability,
	using a filter bank consisting of triangular and trapezoidal band-pass filters.
	We conduct experiments with two methods on two datasets and show that the proposed metrics can highlight the differences between the methods and between the datasets.
	Code is available at \url{https://github.com/shinya7y/UniverseNet}.
}

\section{Introduction}
\label{sec:intro}

As real-world applications of object detection technology continue to advance, the importance of proper object detection evaluation is increasing.
Scale-wise evaluation is particularly important for robotics applications, including safety-critical systems such as self-driving cars and unmanned aerial vehicles,
because the scale of an object is related to the distance to the object and to the time-to-collision~\cite{Caltech_PAMI2012}.

Scale-wise metrics for object detection should satisfy the following requirements:
(1) They should be reliable metrics computed with sufficient numbers of objects.
If the number of objects is small, the variance due to randomness is large and the metric is unreliable.
(2) Object scale divisions should be sufficiently fine-grained.
If the granularity is coarse, it is unclear what scale each metric evaluates.
(3) The design of metrics should correspond to that of object detectors.
If there is no correspondence, it is unclear what design changes should be made to improve the metrics for each scale.
Since object detectors typically use pyramidal structures where feature map resolutions become 1/2 in one level~\cite{SSD_ECCV2016, FPN_CVPR2017}, it will be better to use exponentially-spaced metrics that correspond to such structures.
(4) The target object scale of each metric should be intuitively understandable with a single representative value.
If we need to call each metric by the lower and upper limits of the scale range,
it is inconvenient and could cause confusion with neighboring metrics.

\begin{figure}[H]
	\begin{minipage}[t]{\linewidth}
		\centering
		\includegraphics[width=\linewidth]{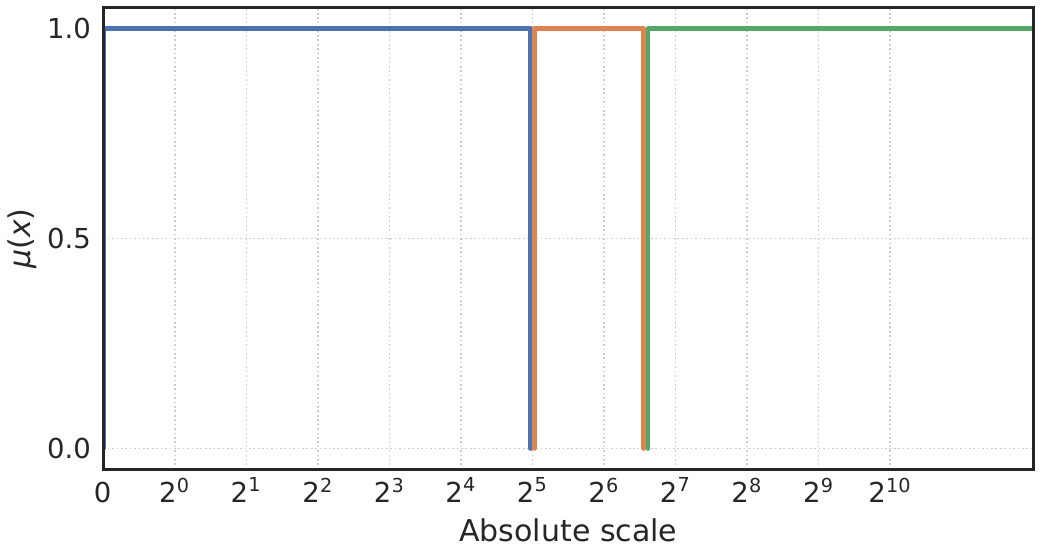}
		\vspace{-5.7mm}
		\subcaption{COCO metrics~\cite{COCO_ECCV2014, cocoapi}}
	\end{minipage}
	\begin{minipage}[t]{\linewidth}
		\centering
		\vspace{3mm}
		\includegraphics[width=\linewidth]{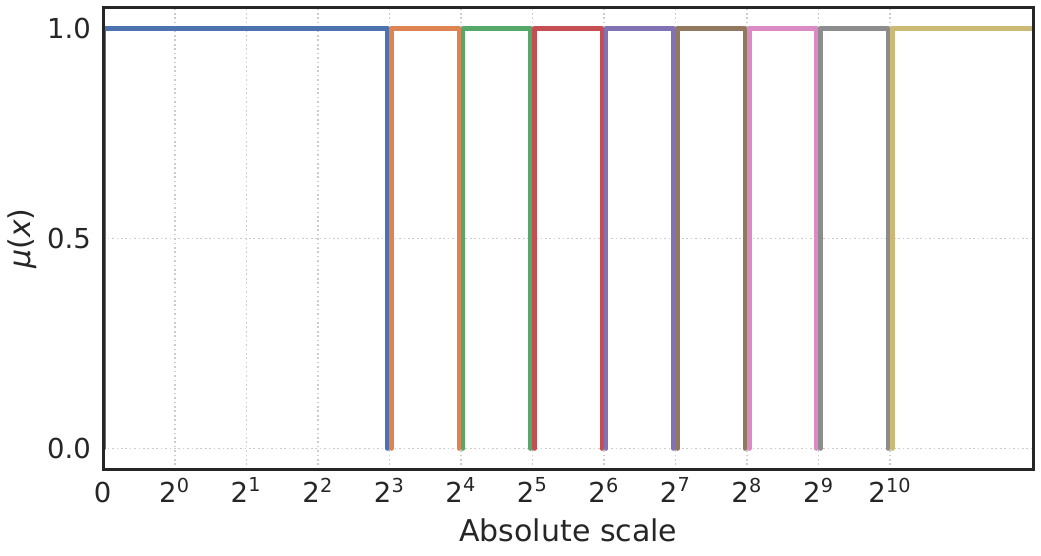}
		\vspace{-5.7mm}
		\subcaption{ASAP~\cite{USB_Shinya_BMVC2022}}
	\end{minipage}
	\begin{minipage}[t]{\linewidth}
		\centering
		\vspace{3mm}
		\includegraphics[width=\linewidth]{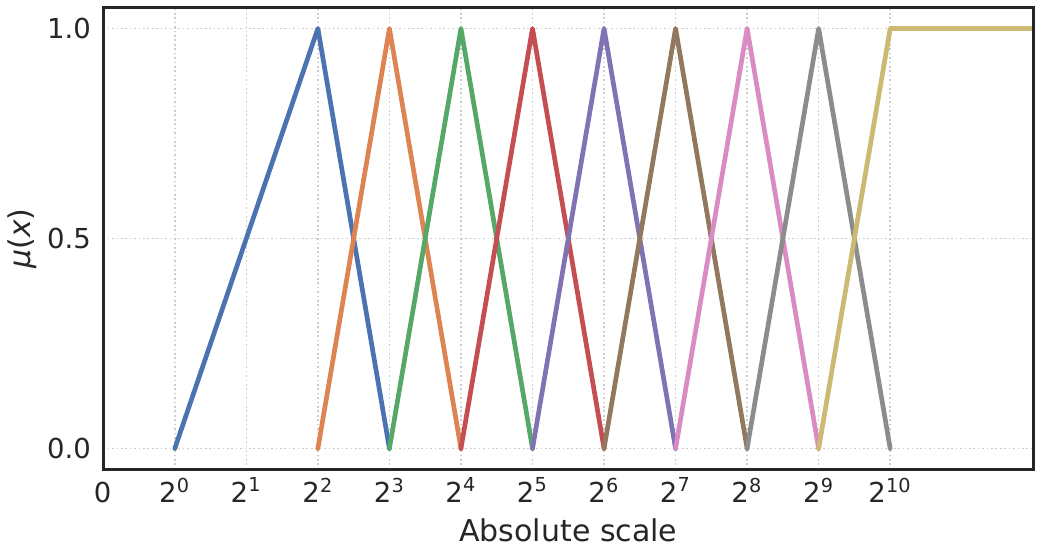}
		\vspace{-5.7mm}
		\subcaption{BandASAP (Ours)}
	\end{minipage}
	\vspace{3mm}
	\caption{
		We rethink the object-scale limitations of scale-wise metrics as band-pass filters.
		(a) Scale-wise metrics of COCO~\cite{COCO_ECCV2014, cocoapi} are too coarse.
		(b) ASAP~\cite{USB_Shinya_BMVC2022} is fine but could be unreliable for some scale ranges.
		(c) We propose BandASAP, which improves the reliability of fine scale-wise metrics.
	}
	\label{fig:filter_bank_comparison}
\end{figure}

Of these requirements, the reliability and fineness of metrics are particularly conflicting.
The popular COCO metrics~\cite{COCO_ECCV2014, cocoapi} use balanced but coarse divisions.
In contrast, the recently proposed ASAP metrics~\cite{USB_Shinya_BMVC2022} are fine but could be unreliable for some scale ranges depending on datasets.

This paper strikes a balance between reliability and fineness by introducing metrics built on the fine ASAP metrics~\cite{USB_Shinya_BMVC2022}.
To this end, we first rethink scale-wise metrics and represent the scale range limitations of existing metrics as rectangular band-pass filters.
Then, by using triangular and trapezoidal band-pass filters, we propose metrics that have peaks at specific object scales
and are computed with more objects than those for the ASAP metrics~\cite{USB_Shinya_BMVC2022}.

Our contributions are as follows:
(1) We propose novel scale-wise metrics for object detection evaluation based on a filter bank consisting of triangular and trapezoidal band-pass filters.
(2) We conduct experiments with two methods on two datasets and show that the proposed metrics can highlight the differences between the methods and between the datasets.

\section{Related Work}
\label{sec:related_work}

There are many coarse scale-wise metrics~\cite{DiagDet_ECCV2012, Caltech_PAMI2012, COCO_ECCV2014, TinyPerson_WACV2020, TIDE_ECCV2020, REVISE_ECCV2020}.
For pedestrian detection, the metrics of the Caltech pedestrian dataset~\cite{Caltech_PAMI2012} use three scale ranges (near, medium, and far) defined by objects' height.
For generic object detection, the COCO metrics~\cite{COCO_ECCV2014, cocoapi} use three scale ranges (small, medium, and large) defined by objects' area.
For tiny person detection, the code of TinyPerson metrics~\cite{TinyPerson_WACV2020} uses five scale ranges (tiny1, tiny2, tiny3, small, and reasonable) defined by objects' absolute scale.
Although these metrics are based on predefined scale values, they take into account the distribution of each corresponding dataset.
However, divisions that are well-balanced for a specific dataset are not necessarily well-balanced for other datasets.
The COCO metrics are particularly popular and are often used as the de facto standard metrics, but they can cause unreliable scale ranges for datasets whose scale distributions are significantly different from that of COCO.
In contrast to the above metrics, Hoiem \etal~\cite{DiagDet_ECCV2012} adopt another approach that explicitly considers the balance of the number of objects.
They use five scale ranges defined by objects' percentile scale for each object category (extra-small: 10\%, small: 20\%, medium: 40\%, large: 20\%, extra-large: 10\%).

For detailed scale-wise analysis, two types of fine scale-wise metrics were recently proposed.
Absolute Scale AP (ASAP) and Relative Scale AP (RSAP)~\cite{USB_Shinya_BMVC2022} use nine scale ranges based on absolute scale and relative scale~\cite{TinyPerson_WACV2020}, respectively.
The work assumes that the metrics are computed on datasets with many objects.
When applied to datasets with few objects or those without objects in a specific scale range, the metrics could be less reliable.

Existing coarse scale-wise metrics do not satisfy the second requirement in \cref{sec:intro},
while existing fine scale-wise metrics may not satisfy the first requirement.
To strike a balance between them, we improve the reliability of fine scale-wise metrics.
We also consider the third and fourth requirements to improve the practicality of metrics.

\section{Rethinking and Proposal for Filter Banks}
\label{sec:proposed_metrics}

In this section, we first rethink the object-scale limitations of scale-wise metrics as band-pass filters, then explain the proposed metrics.
The proposed metrics are based on ASAP~\cite{USB_Shinya_BMVC2022}, allowing for detailed analysis.
Furthermore, they use triangular and trapezoidal band-pass filters to broaden the range of object scales to be evaluated, while making it easy to understand the peak object scale to be focused on.
\Cref{fig:filter_bank_comparison} shows a comparison of the COCO metrics, ASAP, and the proposed metrics called \textit{BandASAP}.

\subsection{Rectangular Filters}

Conventional scale-wise metrics target objects within a specific scale range for evaluation and ignore objects outside the range.
This can be thought of as applying a rectangular filter that weights objects within the range by $1$ and objects outside the range by $0$.
Each rectangular filter is a step function (piecewise constant function) expressed as
\begin{equation} \label{eq:rectangular_filter}
	\mu(x; a, d) =
	\begin{cases}
		0 & (x < a),           \\
		1 & (a \leq x \leq d), \\
		0 & (x > d),
	\end{cases}
\end{equation}
where $\mu(x; a, d)$ is an indicator function,
$x$ denotes its input scale value,
and $a, d$ are constant scale values to control it.
We consider scale values in log space following~\cite{Caltech_PAMI2012}.

In the next subsections,
we will consider other filters that can output real numbers in the interval $[0, 1]$
by considering $\mu(x)$ as the membership function of fuzzy sets~\cite{FuzzySets_Zadeh_1965}.

\subsection{Triangular Filters}

In contrast to existing metrics that use rectangular filter banks, we introduce a triangular filter bank like the Mel filter bank~\cite{MFCC_Davis_1980}.
Each triangular filter is a piecewise linear function expressed as
\begin{equation} \label{eq:triangular_filter}
	\mu(x; a, p, d) =
	\begin{cases}
		0                   & (x < a),        \\
		\frac{x - a}{p - a} & (a \leq x < p), \\
		1                   & (x = p),        \\
		\frac{d - x}{d - p} & (p < x \leq d), \\
		0                   & (x > d),
	\end{cases}
\end{equation}
where $a, p, d$ are constant scale values to control the membership function.
$p$ is the peak scale of the triangular filter, and we calculate it as $p = \frac{a + d}{2}$ by default.

\subsection{Trapezoidal Filters}
\label{sec:proposed_metrics_trapezoidal_filters}

A filter bank with evenly-spaced rectangular or triangular filters could cause some scale ranges with too few objects.
For example, datasets for small object detection may have few large objects~\cite{SOD4SB_MVA2023_challenge},
and the largest or smallest scale range may be narrower than the other scale ranges~\cite{USB_Shinya_BMVC2022}.
In such ranges, scale-wise metrics will be unreliable because they are computed with few specific objects.
In addition, there are cases where triangular filters are inappropriate.
For example, one cannot determine the peak scale for the largest scale range $[512, 100000]$ in pixels.

To address these issues, we adopt trapezoidal filters.
Each trapezoidal filter is a piecewise linear function expressed as
\begin{equation} \label{eq:trapezoidal_filter}
	\mu(x; a, b, c, d) =
	\begin{cases}
		0                   & (x < a),           \\
		\frac{x - a}{b - a} & (a \leq x < b),    \\
		1                   & (b \leq x \leq c), \\
		\frac{d - x}{d - c} & (c < x \leq d),    \\
		0                   & (x > d),
	\end{cases}
\end{equation}
where $a, b, c, d$ are constant scale values to control the membership function.
This function can represent the rectangular filter of \cref{eq:rectangular_filter} when $a = b$ and $c = d$ and the triangular filter of \cref{eq:triangular_filter} when $a < b = c < d$.
We selected the inequality sign to maintain compatibility with the COCO metrics and avoid division by zero.

We use a trapezoidal filter for the largest scale range and triangular filters for the other scale ranges by default.
There is room for improvement in filter selection because which filter should be used for which range is highly dependent on datasets.

\subsection{Weighted Recall and Precision}

The standard computation of Average Precision (AP) does not assume weighting by object scales.
Thus, we replace $\Recall$ and $\Precision$ in the AP calculation of the COCO metrics~\cite{cocoapi} with the following $\wRecall$ and $\wPrecision$:
\begin{equation}
	\wRecall = \frac{\wTP}{\wTP + \wFN},
\end{equation}
\begin{equation}
	\wPrecision = \frac{\wTP}{\wTP + \wFP},
\end{equation}
\begin{equation}
	\wTP = \sum_{i \in \TP} \mu(x_i),
\end{equation}
\begin{equation}
	\wFN = \sum_{j \in \FN} \mu(x_j),
\end{equation}
\begin{equation}
	\wFP = \sum_{k \in \FP} \mu(x_k),
\end{equation}
where $\TP, \FN, \FP$ denote true positives, false negatives, false positives, respectively,
``$\mathrm{w}$'' prefix means weighted versions,
$x_i$ and $x_j$ denote scales of ground truth boxes,
and $x_k$ denotes scales of predicted boxes.

\subsection{Comparison of Scale-Wise Metrics}

Table~\ref{tab:metrics_comparison} shows the comparison with existing metrics
in terms of the four requirements in \cref{sec:intro}.
For simplicity of presentation,
we selected the COCO metrics~\cite{COCO_ECCV2014, cocoapi} and ASAP~\cite{USB_Shinya_BMVC2022}
as representative coarse and fine metrics, respectively.

\noindent
(1) Reliable.
BandASAP typically has twice the scale range of ASAP on a log scale (see \cref{fig:filter_bank_comparison}).
Compared to BandASAP, COCO \APS and \APL have larger scale ranges,
while \APM has a smaller scale range.

\noindent
(2) Fine.
The number of scale ranges is only 3 for the COCO metrics.
The division is so coarse that we cannot distinguish between absolute scales of 100 and 1600~\cite{USB_Shinya_BMVC2022}.
The number of scale ranges is 9 for ASAP and BandASAP,
which is sufficient for analysis on many datasets (see \cite{USB_Shinya_BMVC2022} and \cref{sec:experiments}).
It is also easy to define more scale ranges for higher-resolution datasets,
thanks to the following property.

\noindent
(3) Exponential.
ASAP and BandASAP are both exponentially spaced (evenly spaced on a log scale)
and thus have the advantage discussed in \cref{sec:intro},
unlike the COCO metrics.

\noindent
(4) Intuitive.
Existing metrics do not have peaks that enable intuitive understanding with a single value.
Instead, in Table~\ref{tab:metrics_comparison},
we compare the geometric mean of the infimum and supremum of the scale range closest to COCO \APM
because the geometric mean is more representative than the arithmetic mean for the log-normal distribution of object scales~\cite{Caltech_PAMI2012}.
For the COCO metrics, the geometric mean of the scale range of 32--96 is $32\sqrt{3} \sim 55.4$.
For ASAP, that of 32--64 is $32\sqrt{2} \sim 45.3$.
These values are not intuitively understandable.
In contrast, the value is $64$ for BandASAP.

\begin{table}[t]
	\setlength{\tabcolsep}{2.0mm}
	\caption{
		Comparison of scale-wise metrics.
		$*$: The values represent the number of scale ranges.
		$\dagger$: The values represent the geometric mean of the infimum and supremum of the scale range closest to COCO \APM.
	}
	\vspace{-4.5mm}
	\begin{center}
		\scalebox{0.9}{\begin{tabular}{lcccc}
				\toprule
				Metrics                            & Reliable & Fine$^*$             & Exp. & Intuitive$^\dagger$           \\ \midrule
				COCO~\cite{COCO_ECCV2014, cocoapi} & \cm      & \textcolor{Red}{3}   & \xm  & \textcolor{Red}{$32\sqrt{3}$} \\
				ASAP~\cite{USB_Shinya_BMVC2022}    & \xm      & \textcolor{Green}{9} & \cm  & \textcolor{Red}{$32\sqrt{2}$} \\
				BandASAP (Ours)                    & \cm      & \textcolor{Green}{9} & \cm  & \textcolor{Green}{64}         \\ \bottomrule
			\end{tabular}}
		\label{tab:metrics_comparison}
	\end{center}
\end{table}

\section{Experiments}
\label{sec:experiments}

\subsection{Datasets}

We evaluated the proposed metrics on datasets for generic object detection and small object detection.
For generic object detection, we adopted the standard COCO dataset~\cite{COCO_ECCV2014}.
We used models trained on the \texttt{train2017} split (118,287 images)
and evaluated them on the \texttt{val2017} split (5,000 images).
For small object detection, we adopted the SOD4SB dataset~\cite{SOD4SB_MVA2023_challenge}, a recently released dataset for small object detection for birds.
Bird detection is an important application of small object detection for industrial and environmental reasons~\cite{BirdDataset_Yoshihashi_ICIP2015, DistantBird_Fujii_MVA2021}.
The image resolution is $3,840{\times}2,160$.
We trained models on the \texttt{train} split (8,880 images) and evaluated them on the \texttt{val} split (879 images) and the \texttt{public test} split (9,699 images).

\subsection{Models and Training}

Our code is based on MMDetection~\cite{MMDetection}.
We used GFL~\cite{GFL_NeurIPS2020} for a single-stage detector
and Faster R-CNN~\cite{Faster_R-CNN_NIPS2015} for a multi-stage detector.
The backbone is ResNet-50~\cite{ResNet_CVPR2016}, and the neck is FPN~\cite{FPN_CVPR2017}.
For COCO, we downloaded the pre-trained models from MMDetection, which had been trained with the $1\times$ schedule.

For SOD4SB, we trained models from the COCO pre-trained models.
The image resolution for training and evaluation is $3,840{\times}2,160$.
We used SGD with a momentum of 0.9, a weight decay of $10^{-4}$, and an effective batch size of 16 (with 2 samples per GPU, 1 GPU, and gradient accumulation for 8 steps).
We trained the models for 7 epochs using horizontal flipping for data augmentation.
For GFL~\cite{GFL_NeurIPS2020}, the learning rate is $0.01$ for initial epochs and $0.001$ for the last epoch.
For Faster R-CNN~\cite{Faster_R-CNN_NIPS2015}, the learning rate is $0.02$ for initial epochs and $0.002$ for the last epoch.

We trained the models with an NVIDIA GeForce RTX 3090 GPU.
The GFL model we trained is efficient.
Specifically, the training takes 5 hours with mixed precision.
The latency of inference with a batch size of 1 is 136\MSEC without test-time augmentation (TTA) and 269\MSEC with horizontal-flipping TTA.
The number of parameters is 32\M.

\begin{table}[t]
	\setlength{\tabcolsep}{1.4mm}
	\caption{
		Standard metrics on SOD4SB.
	}
	\vspace{-2mm}
	\begin{center}
		\scalebox{0.9}{\begin{tabular}{lcccc}
				\toprule
				Method                                    & val AP & val AP$_{50}$ & public test AP$_{50}$ & \\ \midrule
				Faster R-CNN~\cite{Faster_R-CNN_NIPS2015} & 36.2   & 68.1          & 54.8                  & \\
				GFL~\cite{GFL_NeurIPS2020}                & 46.5   & 88.6          & 72.4                  & \\
				GFL~\cite{GFL_NeurIPS2020} w/ TTA         & 47.0   & 88.9          & 73.1                  & \\ \bottomrule
			\end{tabular}}
		\label{tab:sod4sb_results}
	\end{center}
\end{table}

\subsection{Results}

\Cref{tab:sod4sb_results} shows standard metrics on SOD4SB.
GFL~\cite{GFL_NeurIPS2020} outperforms Faster R-CNN~\cite{Faster_R-CNN_NIPS2015} by a large margin.
Such a remarkable difference cannot be seen on datasets with many objects of various scales~\cite{USB_Shinya_BMVC2022}.

\begin{figure}[t]
	\centering
	\includegraphics[width=\linewidth]{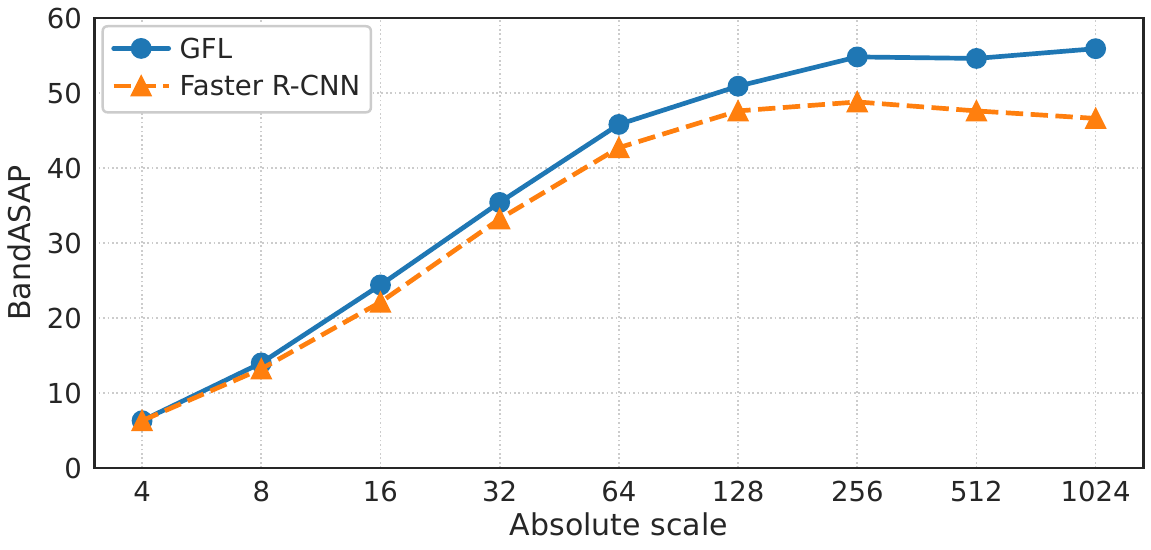}
	\vspace{-8mm}
	\caption{
		BandASAP on COCO.
	}
	\label{fig:band_asap_coco}
\end{figure}

\Cref{fig:band_asap_coco} shows the results of BandASAP on COCO.
They are not particularly strange.
Faster R-CNN~\cite{Faster_R-CNN_NIPS2015} is somewhat weak for large objects
(BandASAP$_{512}$ and BandASAP$_{1024}$ are lower than BandASAP$_{256}$).

\begin{figure}[t]
	\centering
	\includegraphics[width=\linewidth]{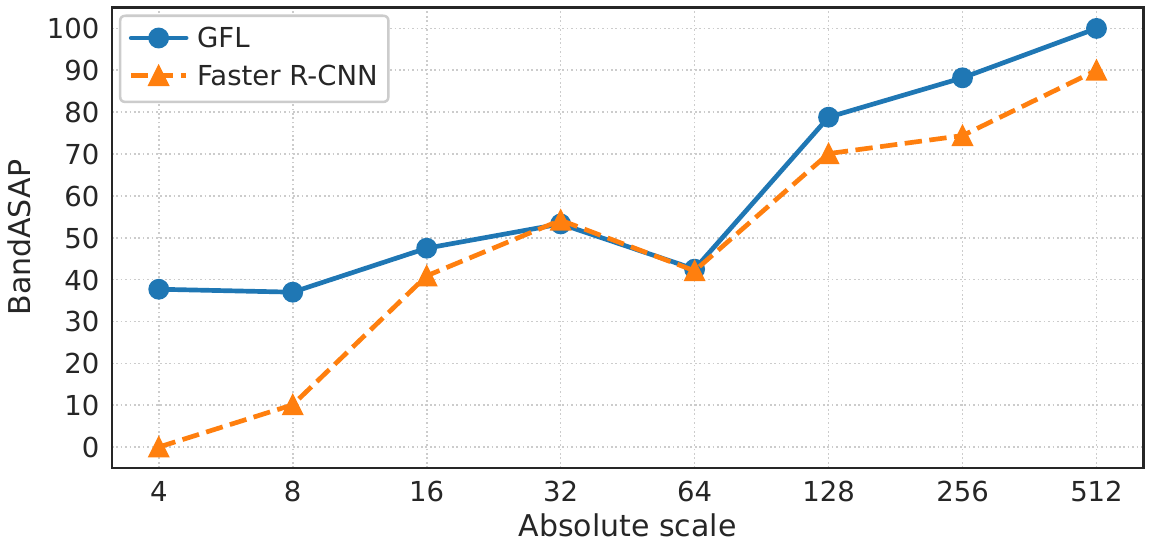}
	\vspace{-8mm}
	\caption{
		BandASAP on SOD4SB.
	}
	\label{fig:band_asap_sod4sb}
\end{figure}

\Cref{fig:band_asap_sod4sb} shows the results of BandASAP on SOD4SB \texttt{val}.
For both methods, BandASAP$_{64}$ is lower than its neighbors, showing valleys.
BandASAP$_{4}$ and BandASAP$_{8}$ of Faster R-CNN~\cite{Faster_R-CNN_NIPS2015} are significantly lower than those of GFL~\cite{GFL_NeurIPS2020}.

\section{Discussions}
\label{sec:discussions}

Why is BandASAP$_{64}$ on SOD4SB low?
This phenomenon is at least partially due to the dataset because it is not happening on COCO.
SOD4SB is a dataset for small object detection, which has many small objects and scarce large ones~\cite{SOD4SB_MVA2023_challenge}.
Thus, a possible cause of low performance in the medium scale range is
a combination of performance degradation due to small object scales and a small number of large objects.
Although small object detection has attracted attention as one of the frontiers of object detection research,
we must not forget that challenges other than small object scales also appear when applying it to real-world problems.

\section{Conclusion}
\label{sec:conclusion}

We proposed BandASAP, a set of scale-wise metrics based on a filter bank consisting of triangular and trapezoidal band-pass filters.
Our experimental results demonstrated the validity of the proposed metrics by highlighting differences between experimental conditions.
There are two limitations to this work.
(1) Although the proposed metrics tell us which scale ranges are inaccurate, they do not provide information on why and what errors are occurring (\eg, the results of Faster R-CNN in \Cref{fig:band_asap_sod4sb}).
Combining our metrics with other analyses, such as TIDE~\cite{TIDE_ECCV2020}, would facilitate actual performance improvement.
(2)
It remains impossible to compute a reliable metric if a widened scale range still has too few objects.
For such cases, a fundamentally different methodology (\eg, generation of evaluation data) will be needed.

\bibliographystyle{ieee_fullname}
\bibliography{usb_mva2023}

\end{document}